# Why teaching resists automation in an AI-inundated era: Human judgment, non-modular work, and the limits of delegation

Songhee Han

Florida State University, Tallahassee, FL 32306, USA

songhee.han@fsu.edu

https://orcid.org/0000-0002-8984-7560

**Abstract:** Debates about artificial intelligence (AI) in education often portray teaching as a modular and procedural job that can increasingly be automated or delegated to technology. This brief communication paper argues that such claims depend on treating teaching as more separable than it is in practice. Drawing on recent literature and empirical studies of large language models and retrieval-augmented generation systems, I argue that although AI can support some bounded functions, instructional work remains difficult to automate in meaningful ways because it is inherently interpretive, relational, and grounded in professional judgment. More fundamentally, teaching and learning are shaped by human cognition, behavior, motivation, and social interaction in ways that cannot be fully specified, predicted, or exhaustively modeled. Tasks that may appear separable in principle derive their instructional value in practice from ongoing contextual interpretation across learners, situations, and relationships. As long as educational practice relies on emergent understanding of human cognition and learning, teaching remains a form of professional work that resists automation. AI may improve access to information and support selected instructional activities, but it does not remove the need for human judgment and relational accountability that effective teaching requires.

## Introduction

Recent advances in generative artificial intelligence (AI), particularly large language models (LLMs), have intensified claims that elements of teaching are becoming automatable. Scholarly debates about the future of educational technology have increasingly focused on whether elements of teaching might be automated or delegated to technology (Rensfeldt & Rahm, 2023; Selwyn, 2019), often extending these task-level possibilities into broader implications for the teaching profession itself. Historically, framing teaching as automatable implicitly treats teaching as a procedural activity whose effectiveness depends on executing predefined steps and whose component tasks can be separated and optimized independently.

This brief communication challenges that assumption. Although the examples and references span multiple educational settings, the argument advanced here is intended to apply broadly across instructional levels and modalities wherever teaching depends on ongoing professional judgment in relation to learners. More specifically, I challenge the idea that teaching can be cleanly decomposed into modular, delegable tasks in ways that would allow AI to take over meaningful portions of instructional work. In this argument, the *teaching job* is not understood as the aggregation of discrete instructional tasks but as a sustained professional practice oriented toward interpreting learners' needs, addressing instructional situations, and supporting understanding over time (Aspelin, 2020; Barnhart & van Es, 2015; Lampert, 2010). This understanding differs from narrower conceptions of teaching as the delivery of information; it emphasizes the interpretive, relational, and judgment-based work through which instruction is made responsive to learners. I argue that instructional work is inherently non-modular, interpretive, and relational. Drawing on empirical and conceptual work on AI in education, including evidence on the limits of LLMs in truthfulness, evaluation, and contextual reasoning, I demonstrate that although AI systems perform well on some bounded, surface-level tasks, they consistently struggle with the kinds of judgment and responsiveness that effective teaching requires.

More fundamentally, teaching and learning are shaped by human cognition, motivation, and social interaction in ways that resist exhaustive specifications or formalization. Instructional decisions are designed in response to evolving learner intentions, motivational misalignment, and context-dependent meanings that emerge through interaction (Laurillard, 2013; Osei & Bjorklund, 2024; Ramsden, 2003), not through stable input–output mappings. As a result, advances in AI do not eliminate the professional judgment at the core of teaching, but instead make its importance more visible.

This paper contributes by arguing that automation narratives mischaracterize teaching as more modular and separable than it is in practice; therefore, they overstate the extent to which meaningful instructional work can be delegated to AI systems. To develop this argument, the paper integrates empirical findings from LLM research with situated learning accounts of human-factor–driven complexity.

**The Core Misconception: Teaching as Modular, Separable Work**

Replacement narratives typically assume that instructional work can be decomposed into stable units that can be optimized independently (e.g., Beard et al., 1973). Conceptually, this assumption reflects a modular view of work familiar in complex systems design, where elements are decomposed and assigned to relatively separable modules according to an overarching architecture or plan (Baldwin & Clark, 2006). Importantly, complexity or context dependence does not by itself rule out modularity. This distinction matters for educational automation because partial delegation to AI depends on the assumption that meaningful aspects of teaching can likewise be separated into sufficiently stable units for independent execution.

The issue in teaching is that apparently discrete tasks such as lesson planning, explanation, feedback, and assessment often derive their instructional value from ongoing interpretation across tasks, learners, timing, and relationships. For example, lesson planning is often framed as selecting content and sequencing activities, feedback as generating comments aligned with a rubric, and grading as assigning scores based on predefined criteria. Under this view, once each component is specified, an automated system could execute these steps at scale. However, this decomposition obscures how instructional decisions are actually made in practice. A lesson plan that works for one group of students may fail for another because of differences in prior knowledge, motivation, classroom dynamics, or emerging misunderstandings (Newton, 2011).

Similarly, feedback that appears appropriate when judged against a rubric may be ineffective if it fails to account for a student's confidence, learning trajectory, or emotional state at the moment it is received (Stevens, 2023). What appears separable in principle is often less separable in practice because the quality of any one instructional act depends on how it is interpreted within a broader pedagogical situation.

This is the central non-modularity problem. The concern is not simply that teaching is contextual or unpredictable, since many complex systems can still be organized modularly. Instead, the concern is that instructional tasks often carry information across one another: assessment informs explanation, feedback shapes future planning, and relational knowledge affects how instructional moves are interpreted by learners. For this reason, teaching cannot be understood simply as a bundle of independent tasks awaiting optimization. Its component activities are interconnected through professional judgment exercised across time and situation (Shavelson & Stern, 1981).

This point is not merely philosophical; it matters because AI systems are strongest when tasks can be specified in advance with stable input–output mappings and evaluable correctness conditions (Ferber et al., 2025; Zhang et al., 2020). If instructional tasks are treated as more separable than they are, automated systems may produce outputs that are fluent, complete, or structurally coherent while still missing the pedagogical meaning of those outputs in context. The core misconception, then, is not simply that teaching is difficult to automate, but that meaningful instructional work can be cleanly separated from the broader interpretive and relational practice of teaching.

## What LLM and Retrieval-Augmented Generation Research Reveals About the Limits of Automation

Having clarified why partial automation depends on task separability, the next question is whether current LLM and RAG-based systems can overcome this limitation in educational contexts. LLMs are probabilistic systems trained on massive corpora of text to generate human-like language by predicting likely sequences of words based on prior context (Brown et al., 2020; Kumar, 2024; Myers et al., 2024). Their apparent fluency often creates the impression that they possess understanding or pedagogical competence. In educational contexts, LLMs have been explored for tasks such as explanation generation, feedback drafting, and conversational support (Abu-Rasheed et al., 2024; Bany Abdelnabi et al., 2025; Xi et al., 2026). However, their outputs are fundamentally shaped by statistical pattern matching, not necessarily grounded comprehension, which introduces well-documented challenges related to factual reliability, hallucination, and overconfidence (Du et al., 2024; Huang et al., 2025; Martino et al., 2023).

Retrieval-augmented generation (RAG) architectures have been proposed as a partial response to these limitations. In a RAG system, a language model first retrieves information from verified, task-specific sources, such as course materials, curated documents, or databases, and then generates a response based on that retrieved content (Lewis et al., 2020; Zhang et al., 2025). For example, instead of answering a student's question from its general training data alone, a RAG system might draw from a course syllabus, assigned readings, or teacher-provided materials before producing a response. By conditioning responses on retrieved content, RAG can improve factual grounding and transparency, making it particularly attractive for educational use where reference fidelity matters.

Recent empirical work on LLMs and RAG systems clarifies both the value and the limits of these tools for education. Studies comparing general-purpose LLMs with reference-grounded RAG systems show that while retrieval substantially improves factual accuracy and source alignment (Abo El-Enen et al., 2025), it does not resolve deeper limitations related to instructional judgment (Yu et al., 2025). In RAG-supported learning environments, improved learning outcomes are often associated less with conversational fluency or factual correctness alone than with learners' own meaning-making work as they grapple with retrieved source materials (Han, 2026). This finding suggests that the educational value of RAG may lie partly in supporting access to relevant information, while the scaffolding, interpretation, and pedagogical judgment needed to make that information educationally meaningful remain with the teacher.

At the same time, empirical evaluations of LLM-based assessment and feedback reveal systematic weaknesses when tasks require interpretation or inference. Research on LLM-as-judge approaches demonstrates that automated evaluators may approximate overall scoring trends (Atkinson & Palma, 2025; Henkel et al., 2025), yet frequently diverge from human judgment in cases involving nuanced reasoning or context-dependent interpretations (Han et al., 2026). High-quality student work that relies on reasonable inference beyond explicit textual correspondence is often undervalued, whereas structurally coherent but shallow responses may be rewarded. These findings indicate that even advanced LLMs remain biased toward literalism and favor surface-level features of performance compared to humans (Mikros, 2025; Yiu et al., 2024).

This line of research clarifies an important limit for educational automation. AI systems, including RAG-enhanced models, perform reliably when tasks can be specified in terms of verifiable correctness, structural coherence, or information retrieval. However, when instructional work requires interpreting why a student responded in a particular way, how feedback is likely to be received, or when to intervene given motivational and emotional conditions, the task is no longer cleanly separable from the broader pedagogical situation. In this sense, empirical LLM and RAG research supports the broader argument of this paper: AI can assist bounded support functions, but it does not remove the need for the professional judgment that effective teaching requires. This limitation is not only a matter of current technical performance. More fundamentally, this limitation reflects why teaching resists modular decomposition in the first place: instructional work depends on interpreting evolving human cognition, motivation, and social meaning under conditions that cannot be fully specified in advance.

## Why "Fully Solving" Teaching Requires Fully Solving the Human Mind

The more fundamental reason teaching resists automation is epistemic: teaching is human-factor driven in ways that cannot be fully formalized (MacLean & Scott, 2007; 2011). Instructional success depends on partially observable internal states, including interest, belonging, perceived competence, anxiety, and trust, as well as on how teachers interpret and respond to social dynamics and contextual constraints (Bardach et al., 2022; Keller, 2009; Tait, 2008). Teachers routinely make high-stakes decisions under uncertainty—when to probe, when to wait, when to challenge, when to protect dignity, and how to balance individual needs with collective classroom functioning (Palmer & Snodgrass Rangel, 2011; Vanlommel & Schildkamp, 2019). The issue is not simply that teaching lacks predefined rules. Contemporary machine learning systems can produce flexible responses without relying on explicit procedural rules. Instead, the educational question is whether those responses support learning in ways that are pedagogically appropriate, relationally accountable, and responsive to learners' changing needs (e.g., Han et al., 2025; Liu et al., 2025).

Educational practice differs from domains where automation can be judged primarily by task completion or output accuracy, because instruction engages with human intentions and behaviors that are sometimes inconsistent, misaligned, or irrational (Hill & Seah, 2023; Park et al., 2023). An AI system may generate an adaptive or context-sensitive response, but successful response generation is not the same as supporting human learning. Regardless of technological sophistication, learners may misjudge their own understanding, pursue counterproductive strategies, or persist in unproductive patterns of engagement. These misalignments often require other humans, particularly teachers, to recognize what is going wrong and to help learners see and interpret it, providing relational support and corrective guidance in ways (Rubach et al., 2023) that exceed what can be assessed through output quality or learners' final observable outcomes. Even highly advanced AI systems may generate responses that appear adaptive, but such responsiveness does not by itself establish that learning is being supported. Teaching concerns both whether a response is correct or useful at a given moment and how that response shapes the learner's ongoing process of understanding, revision, confidence, and engagement. Because of this reason, adaptive response generation should be distinguished from pedagogical support for learning. The latter requires attention to how instructional moves are taken up by learners over time within particular motivational, social, and relational contexts.

This pattern of motivational and behavioral misalignment is well explained by Situated Expectancy–Value Theory (SEVT; Eccles & Wigfield, 2020), which conceptualizes learners' engagement as shaped by subjective expectations for success and task values that are dynamically constructed within specific contexts. From an SEVT perspective, learners' intentions and behaviors are not assumed to be fully rational or internally consistent; instead, motivation fluctuates as perceptions of competence, effort, cost, and value shift across situations and over time. As a result, learners may disengage from tasks they value, persist in tasks they do not understand, or miscalibrate effort relative to actual learning demands. These patterns are not anomalies but theoretically expected features of human learning. Because addressing these misalignments requires interpretive judgment, teachers must recognize when learners' stated intentions diverge from their actual understanding or when confidence, effort, and strategy are miscoordinated. It also requires relational support to help learners recalibrate goals, strategies, and expectations (Martin et al., 2025).

More broadly, learning is not a linear accumulation of information but an emergent process shaped by cognition, motivation, affect, and social interaction. This view aligns with long-standing situated learning perspectives that conceptualize learning as situated, socially mediated, and context-dependent (Lave & Wenger, 1991), far from algorithmic. Students' engagement depends on fluctuating internal states and on interactions with peers, norms, and teacher–student relationships that are neither stable nor fully observable. Teachers continuously interpret these signals, often implicitly, and adjust instruction accordingly. Such judgments are informed by experience, relational knowledge, and situational awareness, not by language fluency or procedural execution.

This complexity is not a temporary gap in scientific understanding. Even with extensive research on cognition and motivation, human learning remains partially unpredictable because it is shaped by lived experience, social meaning, and moment-to-moment interpretation (Brown et al., 1989; Kolb, 2014). Therefore, teaching involves managing uncertainty, not eliminating it. As long as educational practice depends on responding to incomplete and evolving information about learners' internal states and social contexts, instructional effectiveness cannot be guaranteed by automation. This human-factor–origin complexity places a natural ceiling on what instructional work can be mechanized and anchors teaching as a profession grounded in judgment, not computation.

## Implications for Learning Design and Policy

Teaching is non-modular, judgment-centered work. AI systems intended for educational use are best positioned to support tasks where literal correctness and structural consistency are sufficient, such as generating initial drafts of instructional materials, retrieving course-aligned resources, summarizing administrative information, or preparing routine communication templates for teachers to revise. Even in these cases, the instructional value of such support depends on how teachers interpret, adapt, and use it in context. By contrast, tasks that require interpretation, such as evaluating the quality of student reasoning, deciding when to intervene instructionally, responding to disengagement, or mediating conflict, are better kept under teacher control, with AI positioned at most as a supplementary reference. For example, while a RAG system may efficiently surface relevant content or examples for lesson preparation, determining which explanation to emphasize, when to deviate from a plan, or how to respond to confusion depends on contextual cues that cannot be reliably formalized.

Policy initiatives need to move beyond merely providing access to AI tools and instead emphasize professional development that helps educators evaluate where AI support may be useful and where instructional judgment must remain central. Without such guidance, there is a risk that AI will be adopted primarily for speed and convenience, encouraging surface-level use that undermines instructional rigor. As Selwyn (2022) cautions, work that appears routine enough for automation may still carry pedagogical consequences that become visible only in practice. Institutional policies should explicitly protect teacher agency by framing AI as a support infrastructure, not an evaluative authority, particularly in high-stakes contexts such as assessment and feedback. At the system level, governance should prioritize transparency, reference grounding, and clear accountability for AI-assisted decisions, ensuring that responsibility for instructional judgment remains human. Altogether, these learning design and policy choices reinforce the professional role of teachers while allowing AI to reduce selected support burdens in ways that strengthen teaching practice.

## Conclusion

Teaching resists automation not because AI systems are incapable of generating useful outputs, but because teaching is not a set of modular procedures. It is professional judgment exercised under uncertainty, grounded in human motivation, cognition, and social relationships. AI can support teachers by assisting bounded support functions and improving access to information, particularly when grounded architectures such as RAG are used. Yet AI systems continue to struggle with the interpretive and relational labor that makes teaching effective and ethical. For these reasons, efforts to automate teaching encounter principled limits rooted in the need for human judgment and relational accountability. As AI tools rapidly enter classrooms with limited pedagogical consideration, clarifying these limits is increasingly urgent. Doing so is essential for preserving instructional quality and teacher agency.

## Competing Interests

The author reports no potential conflicts of interest.